\documentclass[runningheads]{llncs}
\usepackage[T1]{fontenc}
\usepackage{graphicx}
\usepackage{comment}
\usepackage{import}
\usepackage{booktabs}
\usepackage{subfig}
\usepackage{multirow}
\usepackage{colortbl}
\definecolor{Gray}{gray}{0.9}

\begin{document}
\title{Constrained Reasoning Chains for Enhancing Theory-of-Mind in Large Language Models}
\titlerunning{Constrained Reasoning Chains for Enhancing ToM in LLMs}
%
\author{Zizheng Lin\inst{1}\orcidID{0000-0002-7229-3602} \and
Chunkit Chan\inst{1}\orcidID{0000-0002-1520-4597} \and
Yangqiu Song\inst{1}\orcidID{0000-0002-7818-6090} \and Xin Liu\inst{2}\orcidID{0000-0001-9610-9526}}
\authorrunning{Z. Lin et al.}
%
\institute{HKUST, Hong Kong SAR \and Amazon, USA\\
\email{zlinai@cse.ust.hk}}  
\maketitle              
\begin{abstract}
Theory-of-Mind (ToM) ability possessed by Large Language Models (LLMs) has been shown to be limited.
Most existing methods for improving ToM in LLMs adopt zero-shot prompting, and they face challenges including poor performance in complex ToM reasoning tasks and an inability to handle non-narrative contexts.
We propose a zero-shot prompting method named Constrained Chain-of-ToM (CCoToM) that leverages domain knowledge and the causal relations between ToM dimensions to address these limitations.
Specifically, CCoToM guides LLMs to construct explicit reasoning chains by first prompting LLMs to infer related ToM dimensions (e.g., belief). Afterward, CCoToM prompts LLMs to infer the queried ToM dimension based on the generated related ToM dimensions and corresponding causal relations. 
Additionally, CCoToM adaptively imposes constraints on prompts to introduce inductive biases and improve consistency between ToM dimensions. 
Besides narratives, CCoToM can also handle non-narrative contexts like conversations.
Extensive experiments show that CCoToM consistently outperforms previous state-of-the-art methods by large margins across all LLMs and datasets used.
We also conduct 
in-depth analyses to gain deeper insights into CCoToM.
We have made our code publicly available.\footnote{https://github.com/HKUST-KnowComp/Constrained-Chain-of-ToM}
\keywords{Theory-of-Mind  \and Large Language Models \and Reasoning Chains.}
\end{abstract}

\section{Introduction}
\begin{figure}[!t]
    \vskip -0.5\baselineskip
    \centering
     \scalebox{0.35}{
    \includegraphics{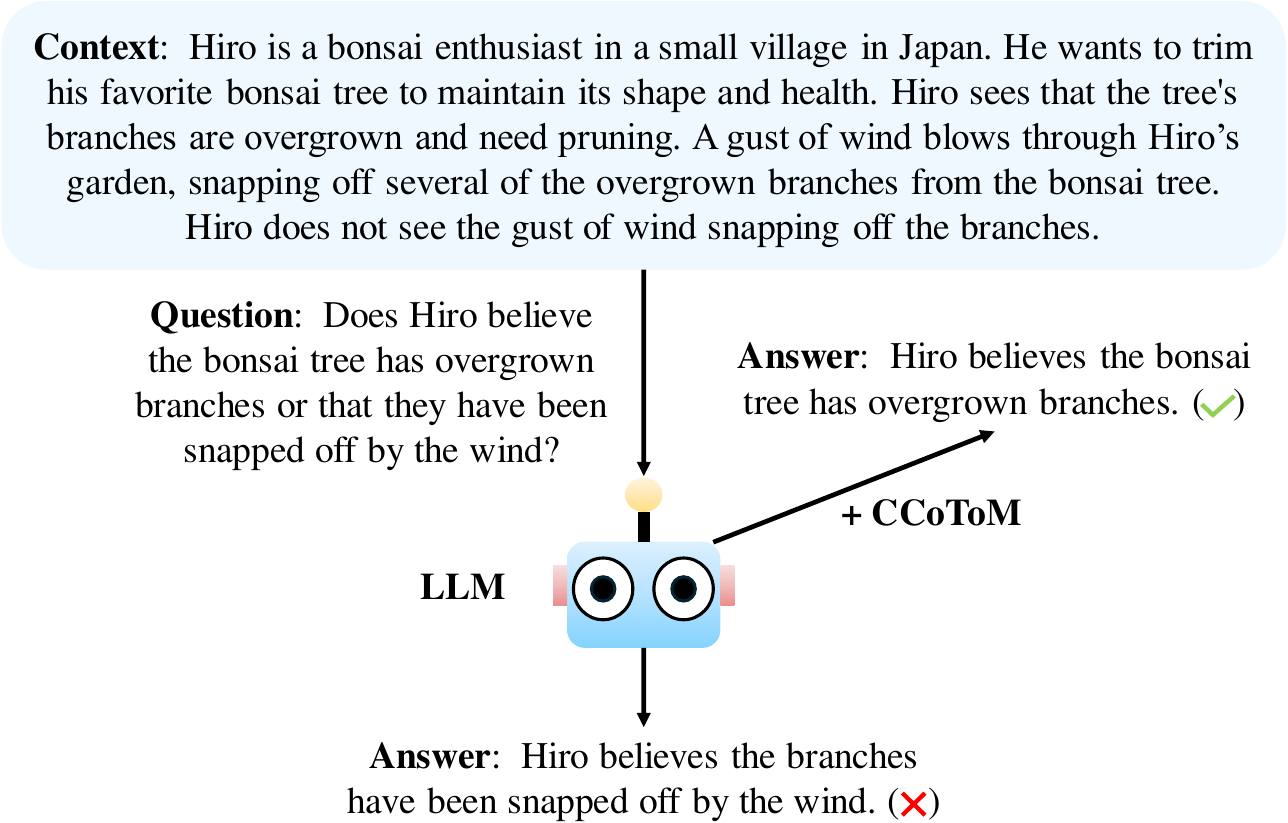}
    }
    \vskip -0.4\baselineskip
    \caption{An example testing the ToM in LLMs. The example is from the Forward Belief task in the BigToM dataset~\cite{bigtom}. 
    Our zero-shot prompting method called CCoToM improves LLMs' ToM reasoning capability.}
    \label{fig:tom_example}
    \vskip -1.5\baselineskip
\end{figure}

Theory of Mind (ToM), introduced as the ability to infer the mental states (i.e., desires, beliefs, and intentions) of others~\cite{DBLP:journals/corr/abs-2404-13627,bigtom,tom_origin}, is crucial for human cognition and social reasoning.   
Recent advancement of Large Language Models (LLM)~\cite{gpt4,openai2022chatgpt,llama2chat70b} attracted abundant research to study the ToM in LLMs~\cite{bigtom,fantom}. An example of testing the ToM in LLMs is shown in Figure \ref{fig:tom_example}, where an LLM is required to infer the belief of an agent (i.e., Hiro in this example). 
Despite being simple for humans, many existing ToM datasets have posed tremendous challenges for LLMs~\cite{bigtom,fantom}.

To improve the ToM in LLMs, supervised fine-tuning is not suitable. 
First, fine-tuning LLMs on existing ToM datasets is prone to overfitting due to the small sizes of these datasets~\cite{symbolictom}. Second, it is costly to fine-tune modern LLMs as they have billion-scale parameters~\cite{openai2022chatgpt,llama2chat70b}. 
Few-shot prompting, on the other hand, is sensitive to the number and content of selected examplars~\cite{DBLP:journals/corr/abs-2302-13539,DBLP:conf/emnlp/ZhangFT22}, thereby failing to robustly improve the ToM in LLMs.
Hence, most existing methods for enhancing the ToM in LLMs adopt zero-shot prompting~\cite{symbolictom,cot,simtom}.
%
Unfortunately, existing zero-shot prompting methods face several challenges. On the one hand, some methods like Chain-of-Thought (CoT)~\cite{cot} and S{\scriptsize IM}T{\scriptsize O}M~\cite{simtom} have poor performance in complex ToM reasoning tasks since they fail to effectively incorporate the causal relations between ToM dimensions\footnote{We use ``ToM dimensions'' to denote percept, belief, desire, and action in the context of ToM. The definitions of these four ToM terms are provided in Section \ref{sec:modeling_causal_relations}.} into LLMs' reasoning processes. 
On the other hand, some methods like S{\scriptsize YMBOLIC}T{\scriptsize O}M~\cite{symbolictom} are difficult to adapt to ToM datasets with non-narrative contexts, and this issue can be mitigated by exploiting domain knowledge~\cite{chang2012structured,DBLP:conf/eacl/LinZS23}.

Inspired by the Belief-Desire-Intention (BDI) model~\cite{bdi}, we propose a zero-shot prompting method named Constrained Chain-of-ToM (CCoToM),  which constructs explicit reasoning chains based on ToM-specific domain knowledge and the causal relations between ToM dimensions to address the aforementioned limitations. Specifically, given a context and a question querying a particular ToM dimension, such as the action of an agent, CCoToM first prompts the underlying LLM to infer the agent's related ToM dimensions like percept, belief, and desire. Afterward, CCoToM prompts the LLM to infer the queried ToM dimension based on the generated related ToM dimensions and corresponding causal relations. In this way, CCoToM guides the LLM to construct explicit reasoning chains throughout the ToM reasoning process, decomposing a complex ToM reasoning task into smaller and simpler sub-tasks. Besides, motivated by the idea of Constrained Modeling~\cite{chang2012structured}, CCoToM adaptively imposes constraints on prompts to introduce inductive biases and improve consistency between ToM dimensions throughout the construction of reasoning chains. 
These constraints are based on domain knowledge and the causal relations, and they are enforced via natural language instructions specifying definitions and dependency. 
Since the domain knowledge and causal relations are agnostic of context type, CCoToM can handle both narrative and non-narrative contexts like conversations. The overview of CCoToM is shown in Figure \ref{fig:method_overview}.

We conduct extensive experiments on two prominent ToM reasoning datasets: BigToM~\cite{bigtom} and FANT{\scriptsize O}M~\cite{fantom}, where the context types of BigToM and FANT{\scriptsize O}M are narratives and conversations respectively.
Experimental results show that CCoToM consistently outperforms previous state-of-the-art (SOTA) methods by large margins across all LLMs and datasets used.
For instance, using GPT-3.5-Turbo~\cite{openai2022chatgpt} as the underlying LLM, CCoToM surpasses previous SOTA methods by up to 41.5 and 18.4 absolute points in accuracy on BigToM and FANT{\scriptsize O}M respectively.
Moreover, to gain deeper insights into CCoToM, we conduct extensive, in-depth analyses, like analyses of related ToM dimensions in reasoning chains.

\section{Modeling Causal Relations between ToM Dimensions}\label{sec:modeling_causal_relations}
\begin{figure}[h]
     \centering
     \vskip -2\baselineskip
     \scalebox{0.4}{
     \includegraphics{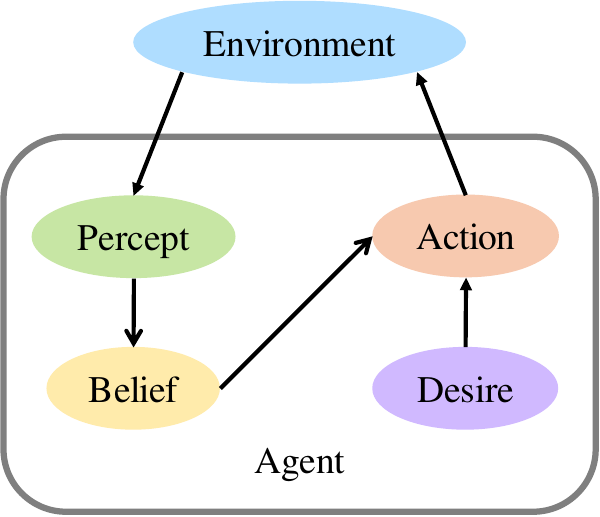}
     }
     \vskip -0.4\baselineskip
     \caption{Overview of the model of causal relations between ToM dimensions.}
     \label{fig:causal_relations}
     \vskip -1.5\baselineskip
 \end{figure}
We describe the model of the causal relations between ToM dimensions of an agent, which is inspired by the BDI model~\cite{bdi}.
The model is shown in Figure \ref{fig:causal_relations}, where ``Percept'' denotes the agent's perception of the environment, ``Belief'' denotes what the agent believes about the environment, ``Desire'' denotes what the agent wants, and ``Action'' denotes the agent's choices with commitment (i.e., intention in the context of ToM) as well as the resulting action in the environment. \textbf{Note that we use ``action'' to represent both intention and action throughout the paper.}  
%
%
Specifically, an agent interacts with the environment by first perceiving information from the environment. The agent's belief is determined by the agent's percept. The agent's belief and desire jointly determine the agent's action in the environment.

\section{Constrained Chain-of-ToM}\label{sec:method}
Based on the model in Section \ref{sec:modeling_causal_relations}, we design a zero-shot prompting method named Constrained Chain-of-ToM (CCoToM).
In this section, we first present an overview of CCoToM. Then we explain how CCoToM guides LLMs to construct reasoning chains. Lastly, we describe how CCoToM imposes constraints.
\subsection{Overview}\label{sec:method:overview}
\begin{figure}[h]
    \vskip -1\baselineskip
    \centering
    \subfloat[Overview of CCoToM for FB task.]{
        \includegraphics[width=0.45\textwidth]{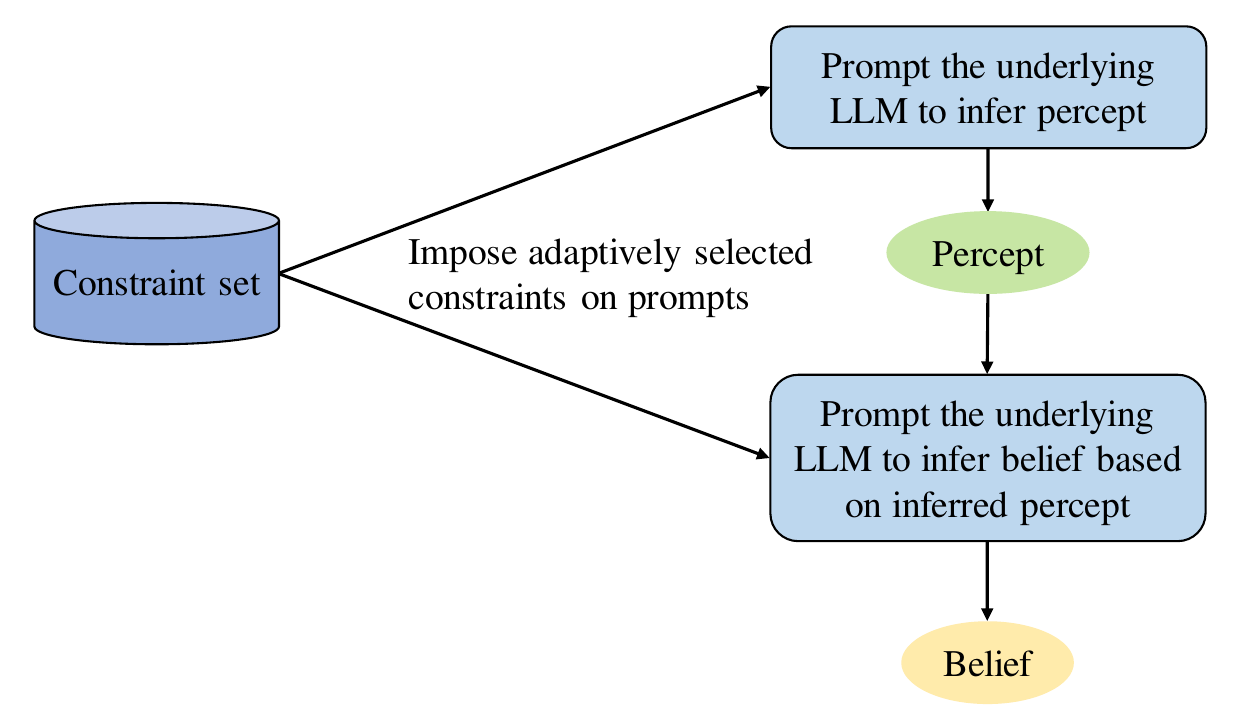}
        \label{fig:overview_forward_belief}
    }
    \hspace{0.05\textwidth}
    \subfloat[Overview of CCoToM for BB task.]{
        \includegraphics[width=0.45\textwidth]{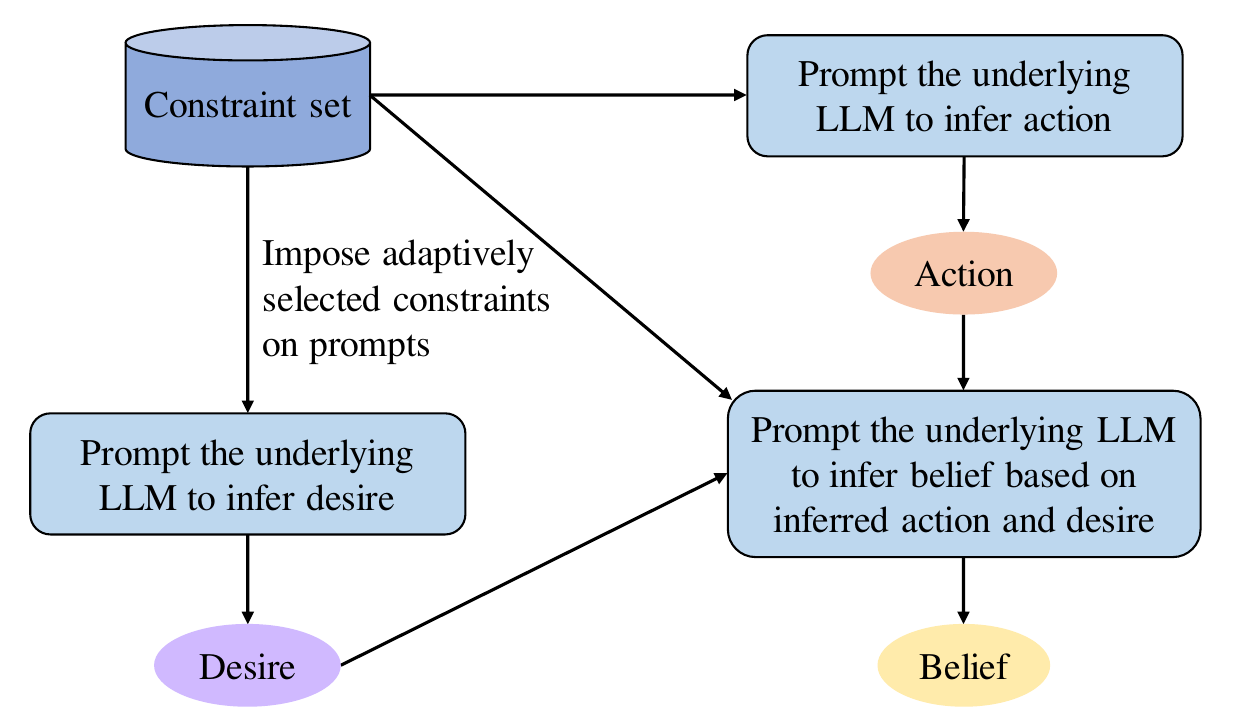}
        \label{fig:overview_backward_belief}
    }    
    \vskip 0.5\baselineskip
    \subfloat[Overview of CCoToM for FA task.]{
        \includegraphics[width=0.55\textwidth]{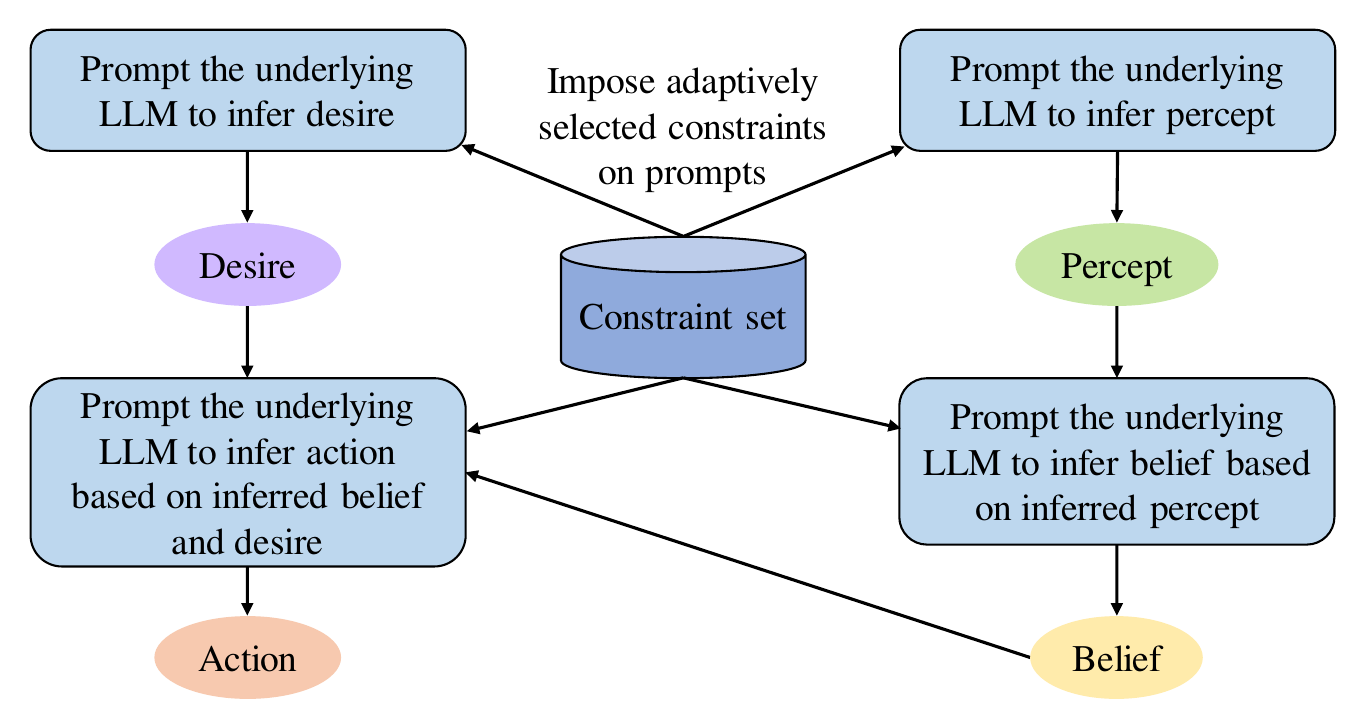}
        \label{fig:overview_forward_action}
    }   
    \vskip -0.5\baselineskip
    \caption{Overview of CCoToM for different types of ToM reasoning tasks. Definitions of the shown task types are introduced in Section \ref{sec:method:overview}. ``Constraint set’’ refers to the collection of CCoToM's constraints.}
    \label{fig:method_overview}
    \vskip -3\baselineskip
\end{figure}

Motivated by the model in Section \ref{sec:modeling_causal_relations}, we categorize ToM reasoning tasks into three types: Forward Belief (FB), Forward Action (FA), and Backward Belief (BB).
In the FB task, LLMs need to infer the belief of an agent given a context containing the agent's percept and desire. 
In the FA task, LLMs need to infer the action of an agent given a context containing the agent's percept and desire. 
In the BB task, LLMs need to infer the belief of an agent given a context containing the agent's action and desire. 
%
%
Overview of CCoToM is shown in Figure \ref{fig:method_overview}. Briefly speaking, CCoToM first prompts the underlying LLM to infer the agent's related ToM dimensions. Thereafter, CCoToM prompts the LLM to infer the queried ToM dimension based on the generated related ToM dimensions and corresponding causal relations.
In this way, CCoToM guides the LLM to construct explicit, ToM-specific reasoning chains, decomposing a complex ToM reasoning task into smaller and simpler sub-tasks.
Additionally, CCoToM adaptively selects constraints from a pre-defined constraint set to impose on prompts via natural language instructions specifying definitions and dependency. 
Moreover, since the domain knowledge and causal relations are agnostic of context types, CCoToM can handle both narrative and non-narrative contexts like conversations.

\subsection{Chain-of-ToM}\label{sec:method:chain-of-tom}
\begin{figure}[h]
    \centering
    \vskip -1.5\baselineskip
    \scalebox{0.5}
    {\includegraphics{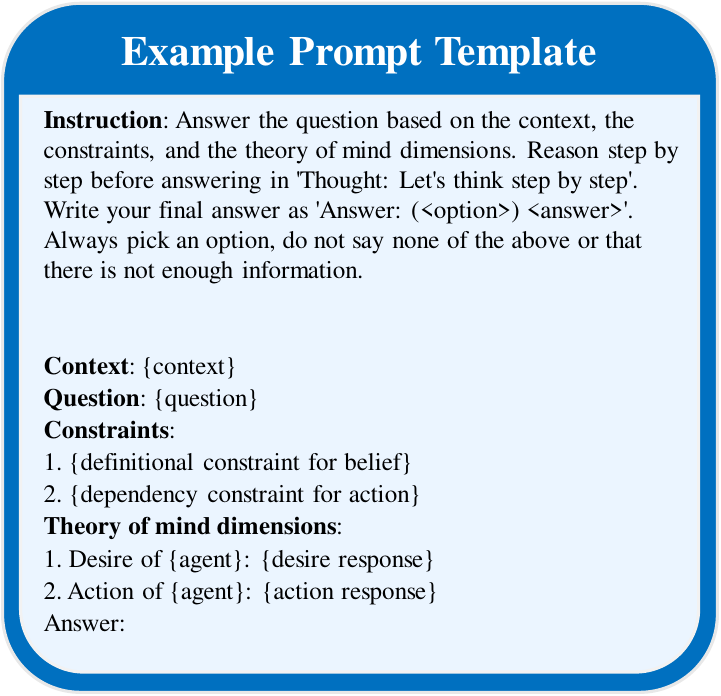}
    }
    \vskip -0.5\baselineskip
    \caption{CCoToM's prompt template for prompting the underlying LLM to infer the belief based on the inferred desire and action for the BB task.
    The constraints shown in the figure are described in Section \ref{sec:method:constraints}. The ``\{agent\}’’ would be replaced by the name of the corresponding agent in the given question (e.g., Hiro in the example of Figure \ref{fig:tom_example}). We explain how CCoToM identifies the name of the agent in Section \ref{sec:method:chain-of-tom}. The ``\{desire response\}'' and ``\{action response\}'' are the inferred desire and action respectively.
}
    \label{fig:example_prompt_template}
    \vskip -2.5\baselineskip
\end{figure}

Every time CCoToM prompts the LLM to infer a certain ToM dimension, \textbf{it only considers the response with the highest likelihood}.\footnote{Regarding closed-source LLMs like GPT-4, we get the response with the highest likelihood by setting the temperature parameter to 0.} 
An example prompt template of CCoToM is shown in Figure \ref{fig:example_prompt_template}. CCoToM identifies the agent's name by prompting the  LLM. For example, given a question querying an agent's belief, CCoToM prompts the LLM to answer the following question: “Whose belief is queried in the given question?”. The other prompt templates are shown in Section \ref{sec:prompt_templates}.
Now we thoroughly describe how CCoToM guides LLMs to construct reasoning chains for different types of ToM reasoning tasks. We adopt the following notations in all equations: $Pe$ for percept, $B$ for belief, $D$ for desire, $A$ for action, $M$ for the underlying LLM, and $C$ for the given context.
Additionally, although the combinations of imposed constraints vary from prompt to prompt,
we still use a single notation $S$ for all combinations of imposed constraints to maintain conciseness.

\subsubsection{Chain-of-ToM for FB Task}\label{sec:method:chain-of-tom:forward_belief}
As shown in Figure \ref{fig:overview_forward_belief}, to tackle the FB Task, CCoToM first prompts the LLM to infer the agent's percept 
based on the context and imposed constraints:
\begin{equation}
    Pe^{*} = argmax_{Pe}P_{M}(Pe|C,S) .
    \label{eq:percept}
\end{equation}
Next, CCoToM prompts the LLM to infer the agent's belief 
based on the inferred percept, the context, and the imposed constraints:
\begin{equation}
    B^{*} = argmax_{B}P_{M}(B|Pe^{*},C,S) .
    \label{eq:percept_to_belief}
\end{equation}

\subsubsection{Chain-of-ToM for FA Task}\label{sec:method:chain-of-tom:forward_action}
As shown in Figure \ref{fig:overview_forward_action}, to tackle the FA Task, CCoToM first prompts the LLM to infer the target agent's percept as in Equation \ref{eq:percept}. Next, CCoToM prompts the LLM to infer the agent's belief as in Equation \ref{eq:percept_to_belief}. Then, CCoToM prompts the LLM to infer the agent's desire 
based on the context and imposed constraints:
\begin{equation}
    D^{*} = argmax_{D}P_{M}(D|C,S) .
    \label{eq:desire}
\end{equation}
Finally, CCoToM prompts the LLM to infer the agent's action 
based on the inferred belief, inferred desire, context, and imposed constraints:
\begin{equation}
    A^{*} = argmax_{A}P_{M}(A|B^{*},D^{*},C,S) .
    \label{eq:belief_desire_to_action}
\end{equation}

\subsubsection{Chain-of-ToM for BB Task}\label{sec:method:chain-of-tom:backward_belief}
As shown in Figure \ref{fig:overview_backward_belief}, to tackle the BB Task, CCoToM first prompts the LLM to infer the target agent's desire, as in Equation \ref{eq:desire}. Next, CCoToM prompts the LLM to infer the agent's action 
based on the context and imposed constraints: 
\begin{equation}
    A^{*} = argmax_{A}P_{M}(A|C,S) .
    \label{eq:action}
\end{equation}
Finally, CCoToM prompts the LLM to infer the agent's belief 
based on the inferred desire, inferred action, context, and imposed constraints:
\begin{equation}
    B^{*} = argmax_{B}P_{M}(B|A^{*},D^{*},C,S) .
    \label{eq:action_desire_to_belief}
\end{equation}

\subsection{Constraints}\label{sec:method:constraints}
CCoToM imposes two types of constraints: definitional and dependency constraints. 
\noindent \textbf{Definitional constraints} describes the semantics of concepts that are crucial in constructing reasoning chains. 
For instance, 
the definitional constraint for belief is ``\textit{Belief of \{agent\} is what \{agent\} believes about the state of the environment.}''.
Definitional constraints introduce inductive biases into the ToM reasoning process since they eliminate ambiguity in reasoning chains' construction by helping LLMs accurately understand the semantics of crucial concepts.
\textbf{Dependency constraints} specify the dependency between ToM dimensions based on the causal relation model in Section \ref{sec:modeling_causal_relations}.
For instance, 
the dependency constraint for action is ``\textit{Action of \{agent\} is determined by the belief of \{agent\} and the desire of \{agent\}.}''.
Dependency constraints improve consistency between ToM dimensions throughout reasoning chains' construction since they guide the LLM to explicitly exploit the cross-dimension dependency. In this way, dependency constraints enhance the constructed reasoning chains' validity with respect to (w.r.t) the causal relation model. 

We designed all constraints based on domain knowledge and the causal relations between ToM dimensions, and the complete set of constraints is shown in Section \ref{sec:complete_constraints} . 
Prior to each time of prompting LLM, CCoToM first selects a combination of constraints based on the task type and ToM dimension to infer. For instance, for BB task, before prompting the LLM to infer belief, CCoToM selects the definitional constraint for belief and the dependency constraint for action. 
Afterward, CCoToM imposes the selected constraints on the prompt by inserting the natural language instructions specifying the constraints into the prompt. 

\section{Experiments}
We first describe the experimental settings. 
We then show and analyze the main results. Additionally, we present extensive in-depth analyses. 
\subsection{Settings}\label{sec:exp:settings}
We use BigToM~\cite{bigtom} and FANT{\scriptsize O}M~\cite{fantom} datasets for evaluation. 
BigToM has three tasks corresponding to three types of ToM reasoning tasks in Section \ref{sec:method:overview}, where each example consists of a narrative context and a Multiple-Choice (MC) question, and \textbf{the evaluation metric is accuracy}.
Each example has three conditions: True Belief (TB), False Belief (FB), and their conjunctive TB\textasciicircum{}FB.\footnote{Definitions of these conditions can be found in Gandhi et al.~\cite{bigtom}.} According to Gandhi et al.~\cite{bigtom}, the accuracy under TB\textasciicircum{}FB condition is the major performance indicator.
The task of FANT{\scriptsize O}M corresponds to FB task.
Each example consists of a conversational context and a question. 
The context can be either ``Short Conversation'' or ``Full Conversation''.
The question can be either ``Choice'' in MC format, or ``Dist'' which requires a free-form response. 
We follow Kim et al.~\cite{fantom} to evaluate each free-form response based on cosine distance computed using SentenceBERT~\cite{sentencebert} embeddings.
\textbf{The evaluation metrics for both types of questions are accuracy}. 
We also follow Kim et al.~\cite{fantom} to calculate the token F1 score for Dist questions.
%
We use the following four SOTA LLMs: GPT-4~\cite{gpt4}, GPT-3.5-Turbo~\cite{openai2022chatgpt} (i.e., ChatGPT), Llama-2 Chat 70B~\cite{llama2chat70b}, and Mistral Instruct 7B~\cite{mistralinstruct7b}. The versions of GPT-4 and GPT-3.5-Turbo used are gpt-4-0613 and gpt-3.5-turbo-0125 respectively. 
We set the temperature parameters of all LLMs to 0 throughout the experiments for reproducibility.\footnote{Since an LLM with the temperature set to 0 always generates the same response for the same input, and our experiments are under zero-shot inference settings with no randomness in the considered tasks, we do not perform repetitive experiments.}
We compare CCoToM with two SOTA zero-shot prompting methods
: CoT~\cite{cot} and S{\scriptsize IM}T{\scriptsize O}M~\cite{simtom}. 
From Section \ref{sec:exp:related_tom_dim} to Section \ref{sec:exp:case_study}, we conduct in-depth analyses using GPT-3.5-Turbo as the underlying LLM and BigToM as the dataset, where we only report TB\textasciicircum{}FB accuracy.

\subsection{Main Results}\label{sec:exp:main_res}
The main results are shown in Table \ref{tab:bigtom_main_res} and Table \ref{tab:fantom_main_res}.
\textbf{Regarding each column of scores, we use bold font to highlight the highest score and largest improvement w.r.t each compared method.} 
We also report CCoToM's average performance gains w.r.t each compared method at the bottom of the tables, where the average is taken across all LLMs.
The results show that CCoToM consistently outperforms previous SOTA methods by large margins across all LLMs and datasets used.

\begin{table*}[!ht]
    \vskip -2\baselineskip
    \caption{Main results on BigToM dataset. In each pair of parentheses, the first number is CCoToM's performance gain w.r.t CoT measured by absolute points in accuracy, and the second number is the one w.r.t S{\scriptsize IM}T{\scriptsize O}M.}
    \vskip 0.2\baselineskip
    \label{tab:bigtom_main_res}
    \centering
    \scalebox{0.38}{
\begin{tabular}{|c|ccc|ccc|ccc|}
\hline
\multirow{2}{*}{Method} & \multicolumn{3}{c|}{Forward Belief} & \multicolumn{3}{c|}{Forward Action} & \multicolumn{3}{c|}{Backward Belief} \\
 & TB & FB & TB\textasciicircum{}FB & TB & FB & TB\textasciicircum{}FB & TB & FB & TB\textasciicircum{}FB \\
\hline
\rowcolor{Gray}
CoT &  &  &  &  &  &  &  &  &  \\
GPT-4 & 97.5 & 96.0 & 93.5 & 98.0 & 87.5 & 85.5 & 82.5 & 58.0 & 41.0 \\
GPT-3.5-Turbo & 90.0 & 94.5 & 85.0 & 91.0 & 35.5 & 32.0 & 78.5 & 64.5 & 47.5 \\
Llama-2 Chat 70B & 83.5 & 90.5 & 79.5 & 87.0 & 34.5 & 30.0 & 74.5 & 53.0 & 33.5 \\
Mistral Instruct 7B & 82.0 & 86.0 & 76.0 & 83.0 & 40.5 & 33.5 & 75.5 & 54.5 & 35.0 \\
\rowcolor{Gray}
S{\scriptsize IM}T{\scriptsize O}M &  &  &  &  &  &  &  &  &  \\
GPT-4 & 95.5 & 97.0 & 95.0 & 93.5 & 92.5 & 87.5 & 77.0 & 53.5 & 35.5 \\
GPT-3.5-Turbo & 87.5 & 95.5 & 86.5 & 83.5 & 58.0 & 48.0 & 74.0 & 58.5 & 41.0 \\
Llama-2 Chat 70B & 84.0 & 89.5 & 82.0 & 80.0 & 55.0 & 44.0 & 73.5 & 48.0 & 30.5 \\
Mistral Instruct 7B & 81.0 & 87.5 & 79.0 & 81.0 & 58.5 & 46.0 & 75.5 & 51.0 & 33.0 \\
\rowcolor{Gray}
CCoToM (Ours) &  &  &  &  &  &  &  &  &  \\
GPT-4 & \textbf{99.0} (+1.5, +3.5) & \textbf{99.0} (+3.0, +2.0) & \textbf{98.0} (+4.5, +3.0) & \textbf{99.5} (+1.5, +6.0) & \textbf{96.5} (+9.0, +4.0) & \textbf{96.0} (+10.5, +8.5) & \textbf{85.0} (+2.5, +8.0) & 66.0 (+8.0, +12.5) & 51.0 (+10.0, +15.5) \\
GPT-3.5-Turbo & 93.5 (+3.5, +6.0) & 98.0 (+3.5, +2.5) & 92.5 (+7.5, +6.0) & 92.0 (+1.0, +8.5) & 77.0 (+\textbf{41.5}, +\textbf{19.0}) & 69.5 (+\textbf{37.5}, +\textbf{21.5}) & 84.0 (+5.5, +\textbf{10.0}) & \textbf{82.0} (+\textbf{17.5}, +\textbf{23.5}) & \textbf{70.0} (+\textbf{22.5}, +\textbf{29.0}) \\
Llama-2 Chat 70B & 90.5 (+\textbf{7.0}, +6.5) & 94.5 (+4.0, +\textbf{5.0}) & 89.0 (+9.5, +7.0) & 89.5 (+2.5, +\textbf{9.5}) & 72.0 (+37.5, +17.0) & 63.5 (+33.5, +19.5) & 80.5 (+6.0, +7.0) & 61.0 (+8.0, +13.0) & 43.0 (+9.5, +12.5) \\
Mistral Instruct 7B & 89.0 (+\textbf{7.0}, +\textbf{8.0}) & 92.0 (+\textbf{6.0}, +4.5) & 87.0 (+\textbf{11.0}, +\textbf{8.0}) & 90.5 (+\textbf{7.5}, +\textbf{9.5}) & 75.0 (+34.5, +16.5) & 66.0 (+32.5, +20.0) & 82.0 (+\textbf{6.5}, +6.5) & 63.5 (+9.0, +12.5) & 45.5 (+10.5, +12.5) \\ 
\hline
Average improvement w.r.t CoT & 4.8 & 4.1 & 8.1 & 3.1 & 30.6 & 28.5 & 5.1 & 10.6 & 13.1 \\
Average improvement w.r.t S{\scriptsize IM}T{\scriptsize O}M & 6.0 & 3.5 & 6.0 & 8.4 & 14.1 & 17.4 & 7.9 & 15.4 & 17.4 \\
\hline
\end{tabular}
}
\vskip -2\baselineskip
\end{table*}
Specifically, regarding BigToM, as shown in Table \ref{tab:bigtom_main_res}, 
the performance gains of CCoToM are up to 11.0, 41.5, and 29.0 in the three tasks respectively, 
showing that CCoToM significantly enhances LLMs' ability to handle all three types of ToM reasoning tasks.
Additionally, the maximum performance gains of combining CCoToM with GPT-4, GPT-3.5-Turbo, Llama-2 Chat 70B, and Mistral Instruct 7B are 15.5, 41.5, 37.5, and 34.5 respectively, which validates CCoToM's effectiveness w.r.t extensive underlying LLMs. 
Moreover, CCoToM's average improvements w.r.t CoT in the TB, FB, and TB\textasciicircum{}FB conditions are up to 5.1, 30.6, and 28.5 respectively, while the corresponding improvements w.r.t S{\scriptsize IM}T{\scriptsize O}M are up to 8.4, 15.4, and 17.4 respectively, which shows that CCoToM substantially enhances the robustness of LLMs' ToM reasoning.
Also, the most notable performance improvements of CCoToM are in the FA task (28.5 and 17.4 average improvement w.r.t CoT and  S{\scriptsize IM}T{\scriptsize O}M respectively), 
which can be due to that solving the FA task requires the most multi-hop ToM reasoning steps, where the benefits from CCoToM's reasoning chains are the most substantial.
In contrast, the least distinct performance improvements of CCoToM are in the Forward Belief task since it is the most simplistic one among the three types of tasks, resulting in the saturated performance of previous SOTA methods.

\begin{table*}[!ht]
    \vskip -1.5\baselineskip
    \caption{Main results on FANT{\scriptsize O}M dataset. In each pair of parentheses, the first number is CCoToM's performance gain w.r.t CoT measured by absolute points in accuracy/F1, and the second number is the one w.r.t S{\scriptsize IM}T{\scriptsize O}M.}
    \vskip 0.2\baselineskip
    \label{tab:fantom_main_res}
    \centering
    \scalebox{0.51}{
\begin{tabular}{|c|ccc|ccc|}
\hline
\multirow{2}{*}{Method} & \multicolumn{3}{c|}{Short Conversation} & \multicolumn{3}{c|}{Full Conversation} \\
 & Choice & Dist. & TokenF1 & Choice & Dist. & TokenF1  \\
\hline
\rowcolor{Gray}
CoT &  &  &  &  &  &  \\
GPT-4 & 74.5 & 38.0 & 47.7 & 67.8 & 37.9 & 46.6 \\
GPT-3.5-Turbo & 45.6 & 29.4 & 40.0 & 40.5 & 27.1 & 37.2 \\
Llama-2 Chat 70B & 43.8 & 27.4 & 29.6 & 38.5 & 25.4 & 28.9 \\
Mistral Instruct 7B & 41.6 & 26.8 & 31.0 & 37.1 & 24.9 & 29.7 \\
\rowcolor{Gray}
S{\scriptsize IM}T{\scriptsize O}M &  &  &  &  &  & \\
GPT-4 & 82.2 & 55.6 & 49.8 & 71.2 & 48.1 & 49.7 \\
GPT-3.5-Turbo & 48.3 & 34.9 & 45.1 & 42.2 & 29.4 & 39.0 \\
Llama-2 Chat 70B & 46.9 & 30.8 & 32.9 & 40.3 & 26.3 & 30.2 \\
Mistral Instruct 7B & 44.2 & 29.1 & 33.2 & 38.5 & 25.5 & 32.3 \\
\rowcolor{Gray}
CCoToM (Ours) &  &  &  &  &  & \\
GPT-4 & \textbf{89.4} (+\textbf{14.9}, +\textbf{7.2}) & \textbf{75.5} (+\textbf{37.5}, +\textbf{19.9}) & \textbf{55} (+7.3, +5.2) & \textbf{79} (+\textbf{11.2}, +\textbf{7.8}) & \textbf{71.7} (+\textbf{33.8}, +\textbf{23.6}) & \textbf{56.2} (+9.6, +6.5) \\
GPT-3.5-Turbo & 51.9 (+6.3, +3.6) & 47.8 (+18.4, +12.9) & 48.4 (+8.4, +3.3) & 45.3 (+4.8, +3.1) & 44.1 (+17.0, +14.7) & 45.8 (+8.6, +6.8) \\
Llama-2 Chat 70B & 50.6 (+6.8, +3.7) & 43.8 (+16.4, +13.0) & 44.6 (+\textbf{15.0}, +11.7) & 44.9 (+6.4, +4.6) & 40.7 (+15.3, +14.4) & 43.1 (+14.2, +\textbf{12.9}) \\
Mistral Instruct 7B & 48.4 (+6.8, +4.2) & 41.4 (+14.6, +12.3) & 45.2 (+14.2, +\textbf{12.0}) & 43.3 (+6.2, +4.8) & 39.6 (+14.7, +14.1) & 44.2 (+\textbf{14.5}, +11.9) \\
\hline
Average improvement w.r.t CoT & 8.7 & 21.7 & 11.2 & 7.2 & 20.2 & 11.7  \\
Average improvement w.r.t S{\scriptsize IM}T{\scriptsize O}M & 4.7 & 14.5 & 8.1 & 5.1 & 16.7 & 9.5 \\
\hline
\end{tabular}
}
\vskip -0.5\baselineskip
\end{table*}
Regarding FANT{\scriptsize O}M, as shown in Table \ref{tab:fantom_main_res}, 
CCoToM outperforms the previous SOTA methods by up to 14.9 and 11.2 absolute points in accuracy in Choice questions of Short and Full Conversation respectively, which shows that CCoToM substantially enhances LLMs' capability of tackling ToM questions with conversational contexts across different context lengths.
Similarly, the performance gains of CCoToM are up to 37.5 and 33.8 in Dist questions of Short and Full Conversation respectively, and the improvements in Token F1 are 15.0 and 14.5 respectively. This demonstrates that CCoToM is adept at tackling ToM questions requiring free-form responses and CCoToM is significantly less susceptible to nonsensical responses~\cite{fantom}.
Additionally, the maximum accuracy improvements of combining CCoToM with GPT-4, GPT-3.5-Turbo, Llama-2 Chat 70B, and Mistral Instruct 7B are 37.5, 18.4, 16.4, and 14.7 respectively, which verifies CCoToM's effectiveness w.r.t extensive underlying LLMs.
Moreover, CCoToM'S average improvements in accuracy w.r.t CoT are up to 8.7 and 21.7 for Choice and Dist questions respectively, while the corresponding improvements w.r.t S{\scriptsize IM}T{\scriptsize O}M are up to 5.1 and 16.7. 
The average accuracy improvements in Dist questions are higher than those in Choice questions, which can be attributed to that solving free-form response questions requires more complicated ToM reasoning than MC questions since the former provides less information (i.e., options are not given), thus making the benefits from CCoToM's reasoning chains to Dist questions more crucial.

\subsection{Analyses of Related ToM Dimensions in Reasoning Chains}\label{sec:exp:related_tom_dim}
\begin{table}[h]
\vskip -3\baselineskip
\caption{Results of analyzing related ToM dimensions in reasoning chains.}
\vskip 0.3\baselineskip
    \label{tab:relate_tom_dim}
    \centering
\scalebox{0.8}{\begin{tabular}{cccc}
\toprule
Task & Configuration & TB\textasciicircum{}FB & $\Delta$ \\
\midrule
\multirow{2}{*}{Forward Belief} & complete method & 92.5 & - \\
 & w/o percept & 87.0 & -5.5 \\
 \midrule
\multirow{4}{*}{Forward Action} & complete method & 69.5 & - \\
 & w/o percept & 63.0 & -6.5 \\
 & w/o belief & 51.0 & -18.5 \\
 & w/o desire & 67.5 & -2.0 \\
 \midrule
\multirow{3}{*}{Backward Belief} & complete method & 70.0 & - \\
 & w/o desire & 66.5 & -3.5 \\
 & w/o action & 53.0 & -17.0 \\
 \bottomrule
\end{tabular}}

\vskip -1.5\baselineskip
\end{table}
%
We analyze the effects of each inferred related ToM dimension. 
We ablate each related ToM dimension while keeping the corresponding constraints.
The results are presented in Table \ref{tab:relate_tom_dim} and we have the following observations: 
(1) Removing any related ToM dimension would cause a drop in accuracy, indicating that every 
dimension contributes to the CCoToM's performance.
(2) The drops in the accuracy range from 2.0 to 18.5, where the most noticeable ones are the belief and action dimensions in FA and BB tasks, respectively. Such a phenomenon is reasonable since the belief of the target agent is the major factor for differentiating options in each question of the FA task in BigToM, and similarly for action in the BB task in BigToM.
Regarding the FB task in BigToM, removing the percept dimension results in a relatively less distinct performance drop, which is due to the task's simplicity causing the saturated performance of LLMs.

\subsection{Analyses of Constraints}\label{sec:exp:constraints}
\begin{figure}[h]
\vskip -1.5\baselineskip
    \centering
    \scalebox{0.23}{
    \includegraphics{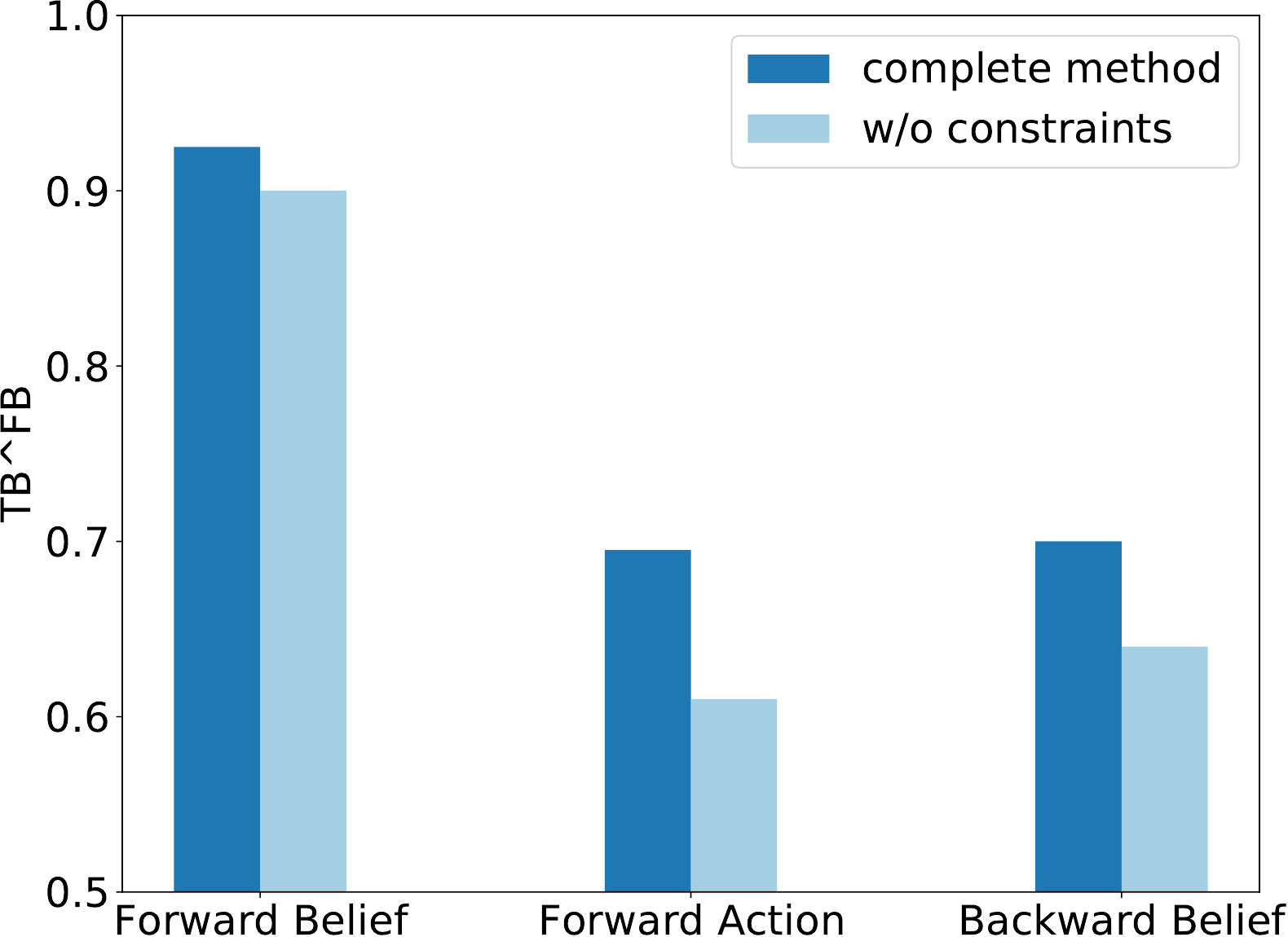}
    }
    \vskip -0.2\baselineskip
    \caption{Results of analyzing constraints.}
    \label{fig:analyses_of_constraints}
    \vskip -1.5\baselineskip
\end{figure}
We analyze the effects of constraints 
on different types of ToM reasoning tasks.
The results are shown in Figure \ref{fig:analyses_of_constraints} and we have the following observations: (1) Removing constraints always leads to a drop in accuracy for all three types of tasks, 
showing the significance of constraints in constructing explicit, ToM-specific reasoning chains. 
(2) The largest performance drop is from the FA task. The reason is that CCoToM for this task involves the most related ToM dimensions, and the inductive biases and consistency between ToM dimensions introduced by constraints are crucial in correctly inferring the related ToM dimensions, thus making the role of constraints in this task the most significant.
In contrast, CCoToM for the FB task involves the least related  ToM dimension, causing the smallest performance drop in this task.

\subsection{Comparison between One-Step and Multi-Step Prompting}\label{sec:exp:one-step_multi-step_prompting}
\begin{figure}[h]
    \centering
    \vskip -0.5\baselineskip
    \scalebox{0.23}
    {
    \includegraphics{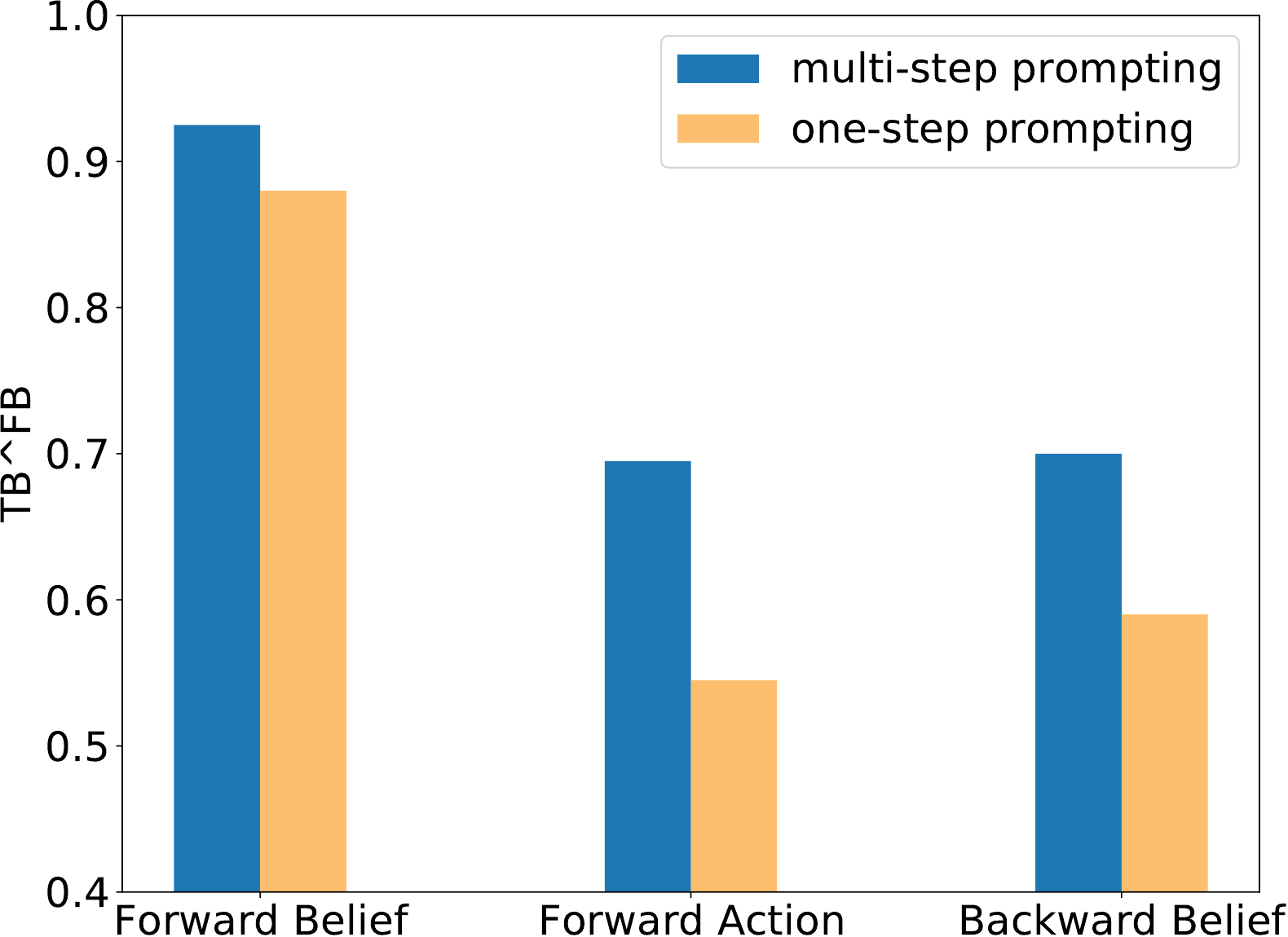}
    }
    
    \vskip -0.2\baselineskip
    
    \caption{Results of comparison between one-step and multi-step prompting.}
    \label{fig:one-step_prompting_vs_multi-step_prompting}
    \vskip -1.5\baselineskip
\end{figure}

We compare CCoToM with a variation of it that uses one-step prompting 
to construct reasoning chains. 
In contrast to CCoToM which uses multi-step prompting, 
the variant condenses all the steps into one prompt by adding instructions that directly specify how to construct reasoning chains into the prompt.
Specifically, the variant adds natural language instructions that directly specify how to construct reasoning chains into the prompt. Take the FB task as an example, such instructions would be ``\textit{First, infer the percept of {agent}. Next, answer the question based on the inferred percept of {agent}.}''. Additionally, the variant includes all related constraints in the prompt.
The results are shown in Figure \ref{fig:one-step_prompting_vs_multi-step_prompting} and we have the following observations: (1) CCoToM consistently outperforms its 
variant across all 
tasks. 
The reason can be that in CCoToM the LLM only needs to tackle one sub-task at each prompting step, whereas in the variant the LLM needs to tackle all sub-tasks in one prompting step, thus making it more difficult for the LLM to accurately construct the reasoning chains.
The reason can be that it is more difficult to tackle all sub-tasks in one prompting step.
(2) The largest performance drop is from the FA task. This can be due to that the decomposition of the FA task generates the most sub-tasks.
This can be due to that the decomposition of the FA task generates more sub-tasks than the decomposition of the other two tasks, which further increases the difficulty of constructing accurate reasoning chains in one prompting step.
In contrast, the smallest performance drop is from the FB task since the decomposition of this task generates the least sub-tasks.

\subsection{Case Study}\label{sec:exp:case_study}

\begin{figure}[h]
    \vskip -2\baselineskip
    \centering 
    \scalebox{0.45}
    {
    \includegraphics{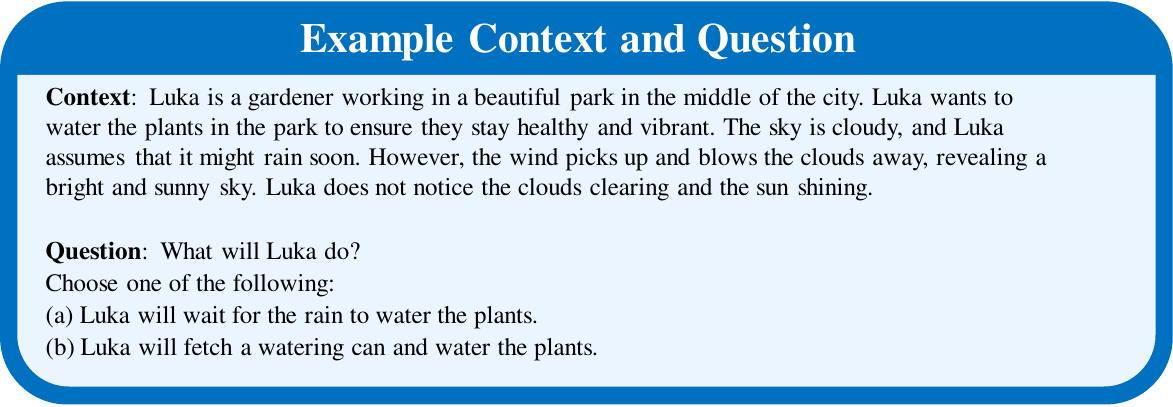}
    }
    \vskip -0.5\baselineskip
    \caption{Example context and question used in the case study.
}
    \label{fig:case_study_example_context_question}
    \vskip -1.3\baselineskip
\end{figure}

\begin{figure}[h]
\vskip -1\baselineskip
    \centering
    \scalebox{0.45}
    {
    \includegraphics{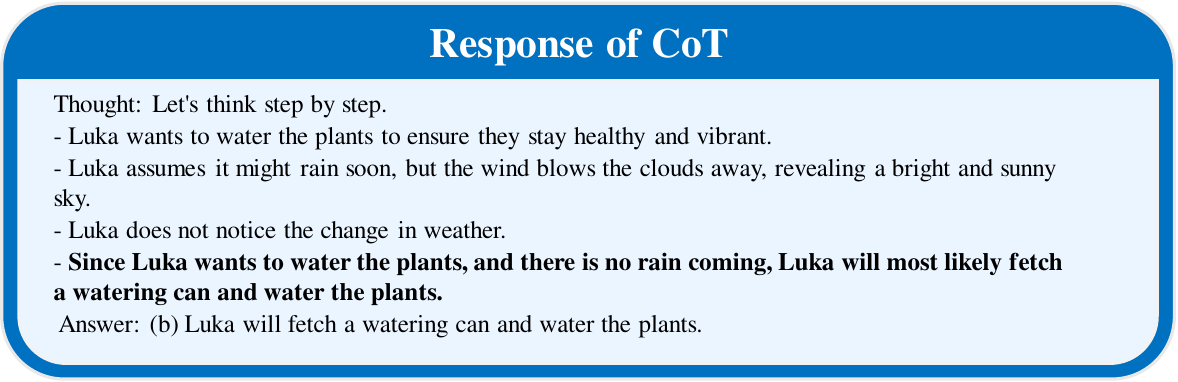}
    }
    \vskip -0.5\baselineskip
    \caption{Response generated by CoT given the example context and question in Figure \ref{fig:case_study_example_context_question}.
}
    \label{fig:case_study_cot_response}
    \vskip -2\baselineskip
    \vspace{-0.2cm}
\end{figure}

\begin{figure}[h]
\vskip 1.5\baselineskip
    \centering
    \scalebox{0.45}
    {
    \includegraphics{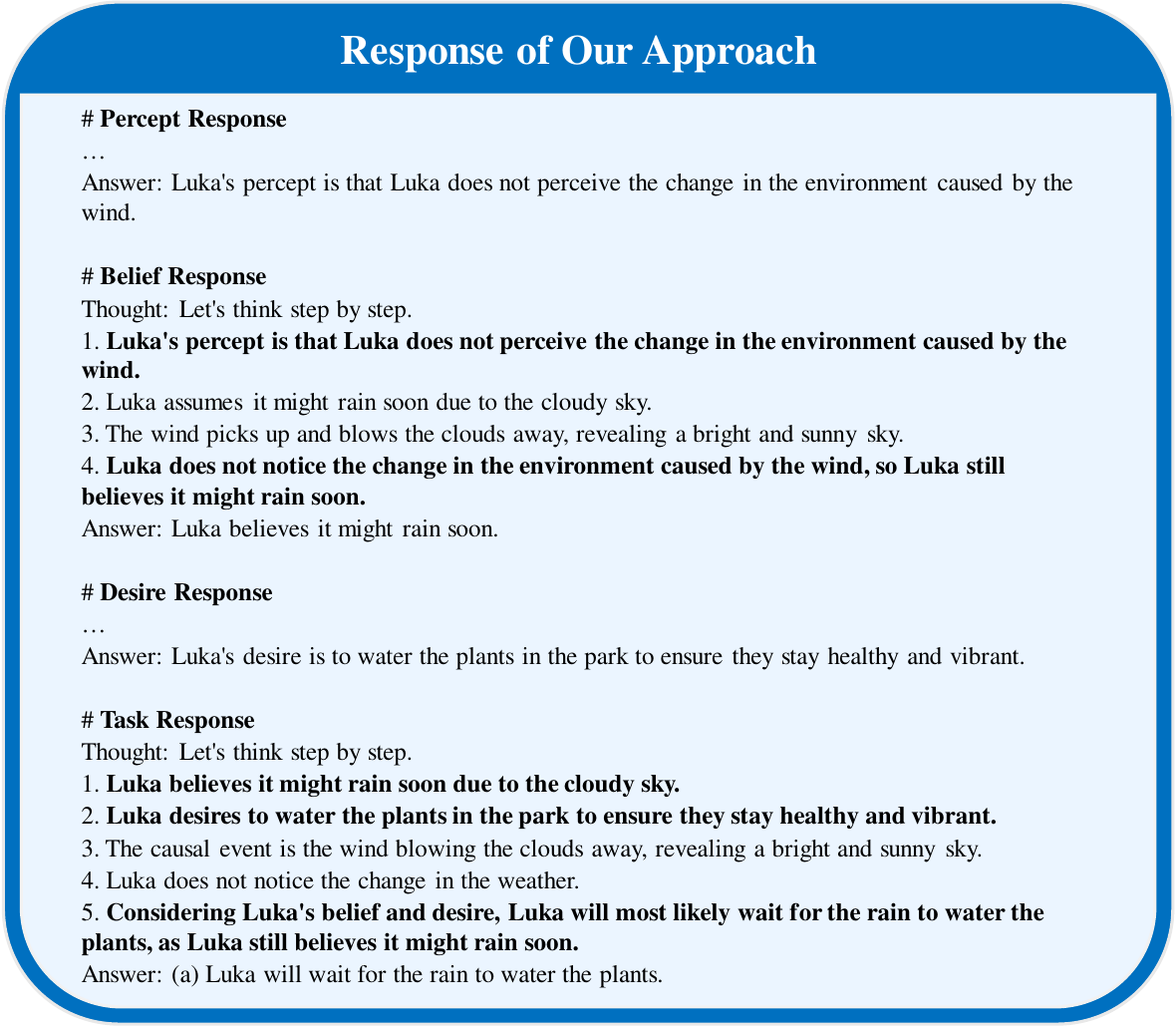}
    }
    \vskip -0.5\baselineskip
    \vspace{-0.2cm}
    \caption{Response generated by our approach given the example context and question in Figure \ref{fig:case_study_example_context_question}, including the response about related ToM dimensions (e.g., belief response) and the task response. We omit some parts of the responses due to space limitation. The omitted parts are replaced by ``…’’.
}
    \label{fig:case_study_response_of_our_approach}
    \vskip -1.6\baselineskip
\end{figure}

In this section, we conduct a case study on the responses of CoT and CCoToM to gain deeper insights into how CCoToM enhances the ToM reasoning capability of LLMs.
The example used in the case study is from the Forward Action task in BigToM and is shown in Figure \ref{fig:case_study_example_context_question}, where the correct answer is option (a).
The responses of CoT and CCoToM are shown in Figure \ref{fig:case_study_cot_response} and Figure \ref{fig:case_study_response_of_our_approach} respectively, where some critical parts of the responses are highlighted in bold font.
In Figure \ref{fig:case_study_response_of_our_approach}, the ``\textbf{Belief Response}'' refers to the response generated by the underlying LLM when it is prompted by CCoToM to infer the belief of the target agent, which is Luka in the example, similarly for the ``\textbf{Percept Response}'' and ``\textbf{Desire Response}'', whereas the ``\textbf{Task Response}'' refers to the response for answering the question in Figure \ref{fig:case_study_example_context_question}.

As shown in Figure \ref{fig:case_study_cot_response}, the LLM combined with CoT chooses the incorrect option since it fails to effectively incorporate the causal relation between percept and belief and the causal relation between belief and action.
In contrast, as shown in Figure \ref{fig:case_study_response_of_our_approach}, CCoToM helps the underlying LLM choose the correct option via leveraging domain knowledge and the causal relations between ToM dimensions to guide the LLM to construct the explicit, ToM-specific reasoning chains.
Specifically, CCoToM first prompts the LLM to infer the percept of Luka, and the inferred percept is ``\textit{Luka does not perceive the change in the environment caused by the wind}'', consistent with the context.
Next, CCoToM prompts the LLM to infer the belief of Luka based on the inferred percept, and the inferred belief is ``\textit{Luka still believes it might rain soon}'', consistent with the context. 
Thereafter, CCoToM prompts the LLM to infer the desire of Luka, and the inferred desire is ``\textit{water the plants in the park}'', consistent with the context.
Finally, CCoToM prompts the LLM to answer the question (i.e., inferring the action of Luka) based on the inferred belief and desire, which guides the LLM to successfully choose the correct option:``\textit{Luka will wait for the rain to water the plants.}''

\section{Related Works}
\textbf{Theory-of-Mind.} One major field of related works is \textbf{Theory-of-Mind} (ToM)~\cite{DBLP:journals/corr/abs-2404-13627,bigtom,tom_origin,DBLP:conf/icml/RabinowitzPSZEB18}, which is crucial in many scenarios involving human cognition and social reasoning~\cite{yim2024evaluating,ijcai2023p330,baron1997mindblindness,DBLP:conf/ro-man/FavierSA23,gopnik1992child,DBLP:conf/aaai/MaoLNH24,DBLP:journals/ai/PereiraPS16}.
Previous studies 
examined the ToM in LLMs 
via various text-based assessments in narrative or conversation scenarios~\cite{DBLP:journals/corr/abs-2404-13627,bigtom,fantom,DBLP:conf/emnlp/LeBN19,DBLP:conf/emnlp/MaSPC23,DBLP:conf/emnlp/Sap0FC22,DBLP:conf/eacl/ShapiraLAZCGSS24}, revealing that LLMs struggle with these tasks~\cite{bigtom,fantom}. 
Most existing approaches for improving ToM in LLMs adopt zero-shot prompting, and they face challenges including poor performance in complex ToM reasoning tasks~\cite{cot,simtom} and an inability to handle non-narrative contexts~\cite{symbolictom}.
%
Another major field of related works is \textbf{Constrained Modeling} (CM), which is a crucial paradigm in machine learning and natural language processing, aiming to integrate prior knowledge into models via constraints~\cite{DBLP:conf/kdd/BasuBM04,ChangSR13,chang2012structured,DeutschUR19,DBLP:conf/aaai/FaghihiNZM0UWPR23,DBLP:journals/jmlr/GanchevGGT10,DBLP:conf/cikm/LinKWBSZY21}. 
In zero-shot scenarios, the knowledge incorporated by CM can offer valuable indirect supervision to models~\cite{DBLP:conf/eacl/LinZS23}. 
However, previous works in the ToM tasks have not explored the CM method to enhance ToM in LLMs. 

\textbf{Large Language Models.} 
With the increase of the computational capabilities and accessibility of vast text corpora, instruction following large language models (LLMs)~\cite{DBLP:journals/corr/abs-2303-08774,openai2022chatgpt,DBLP:conf/emnlp/JiangCCW23} have attained exceptional performance across a diverse array of natural language processing benchmarks in few-shot and even zero-shot learning~\cite{DBLP:journals/corr/abs-2303-12712,DBLP:journals/corr/abs-2302-04023,DBLP:conf/eacl/ChanCWJFLS24,DBLP:conf/acl/0001FLS0XWBLJCS24,DBLP:journals/corr/abs-2309-08303}. Nonetheless, there exist persistent challenges that remain unattended, such as theory of mind reasoning~\cite{DBLP:journals/corr/abs-2404-13627,bigtom,fantom}, analogies reasoning~\cite{DBLP:conf/emnlp/ChengQCFWCRGZSZ23}, text-to-table generation~\cite{DBLP:journals/corr/abs-2404-14215}, complex mathematical reasoning~\cite{DBLP:journals/corr/abs-2301-13867}, implicit discourse relation recognition~\cite{DBLP:conf/acl/ChanLCLSWS23}, address associated ethical concerns, and privacy preservation~\cite{DBLP:journals/corr/abs-2310-10383,DBLP:journals/corr/abs-2302-00539,DBLP:conf/acl/0003GLFH0CYYS24,DBLP:journals/corr/abs-2405-07667}. Therefore, by utilizing these instructions-following LLMs (e.g., GPT-4), we explore the constrained modeling method to enhance the ToM capability of LLMs.

\section{Conclusion and Future Work}
We propose CCoToM, a zero-shot prompting approach that constructs explicit reasoning chains based on domain knowledge and causal relations to strengthen ToM in LLMs.
Besides narratives, CCoToM can also handle non-narrative contexts like conversations.
Extensive experiments on two prominent ToM reasoning datasets show that CCoToM significantly outperforms previous SOTA methods. We also conduct in-depth analyses to gain deeper insights into CCoToM.
In the future, we plan to extend CCoToM to handle more input modalities besides text (e.g., images), or incorporate the grounding method to retrieve mental state from the knowledge graph~\cite{DBLP:conf/coling/JiayangQC0SZ24}.

\subsubsection{Acknowledgements} The authors of this paper were supported by the NSFC Fund (U20B2053) from the NSFC of China, the RIF (R6020-19 and R6021-20) and the GRF (16211520 and 16205322) from RGC of Hong Kong.

\appendix
\section{Prompt Templates}\label{sec:prompt_templates}
In this section, we present all prompt templates for BigToM~\cite{bigtom} and FANT{\scriptsize O}M~\cite{fantom} datasets respectively, except the one in Section 3.2 of the main content.

\subsection{BigToM}\label{sec:prompt_templates:bigtom}
We present the prompt templates for inferring related ToM dimensions (e.g., belief in Forward Action task) and inferring queried ToM dimensions (e.g., belief in Forward Belief task) in Section \ref{sec:prompt_templates:bigtom:related} and Section \ref{sec:prompt_templates:bigtom:quried} respectively.

\subsubsection{Inferring Related ToM Dimensions}\label{sec:prompt_templates:bigtom:related}
\begin{figure}[h]
    \centering
    \scalebox{0.5}{
    \includegraphics{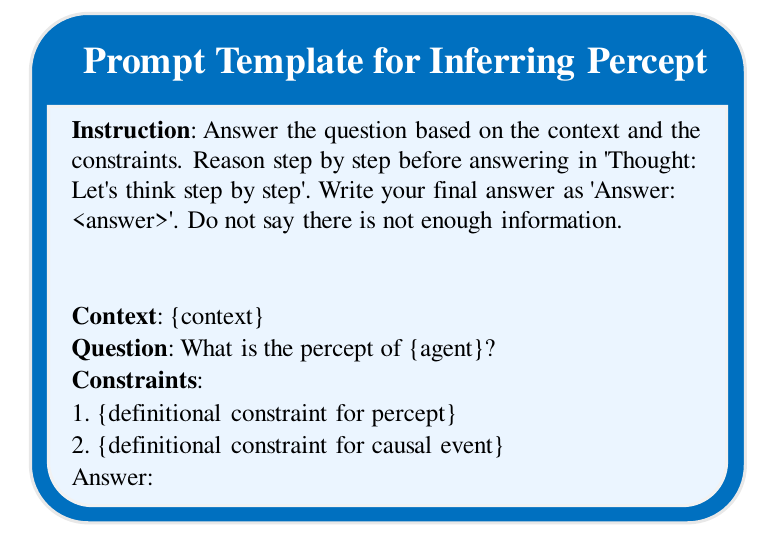}
    }
    \caption{The prompt template for prompting the underlying LLM to infer percept for Forward Belief and Forward Action tasks in BigToM. The constraints shown in the figure are described in Section \ref{sec:complete_constraints}. ``Causal event’’ is the event that changes the state of the environment in the corresponding context.
}
    \label{fig:bigtom_percept_prompt}
\end{figure}

\begin{figure}[h]
    \centering
    \scalebox{0.5}{
    \includegraphics{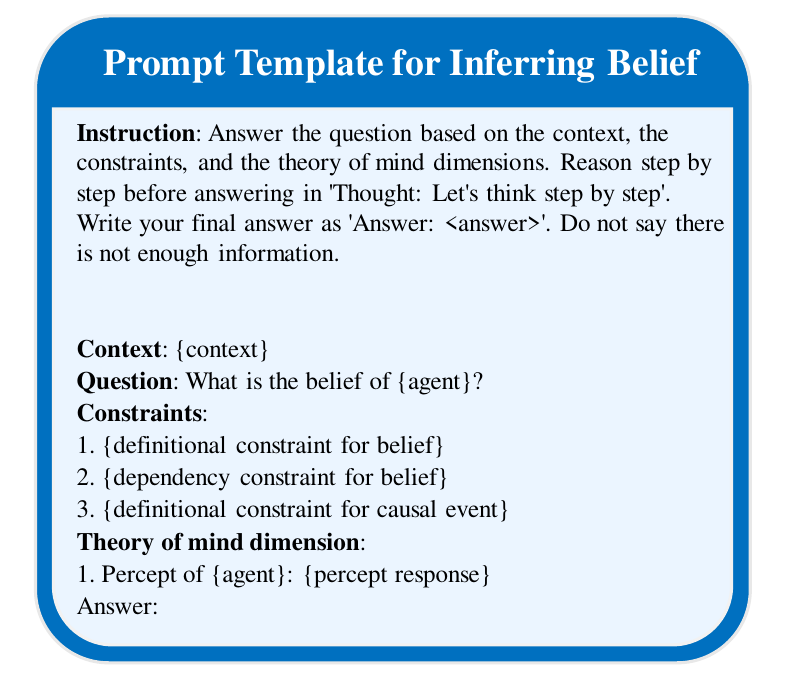}
    }
    \caption{The prompt template for prompting the underlying LLM to infer belief for Forward Action tasks in BigToM. The constraints shown in the figure are described in Section \ref{sec:complete_constraints}. ``\{percept response\}’’ refers to the inferred percept.
}
    \label{fig:bigtom_related_belief_prompt}
\end{figure}

\begin{figure}[h]
    \centering
    \scalebox{0.5}{
    \includegraphics{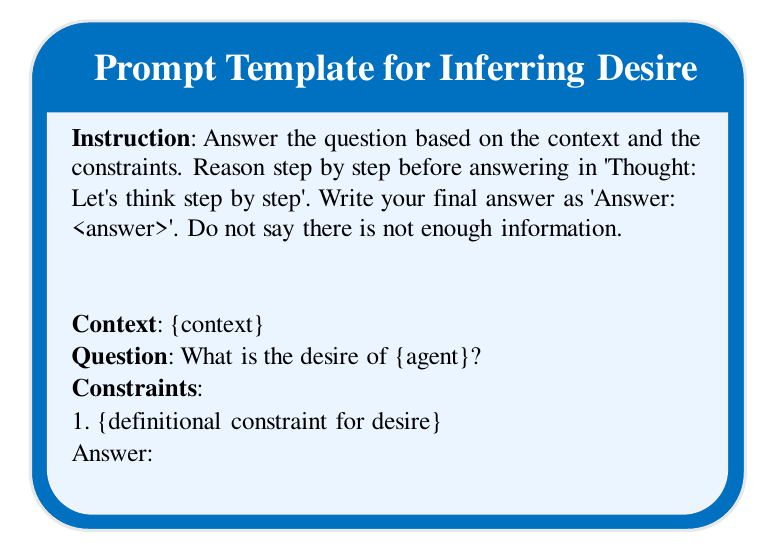}
    }
    \caption{The prompt template for prompting the underlying LLM to infer desire for Backward Belief and Forward Action tasks in BigToM. The constraints shown in the figure are described in Section \ref{sec:complete_constraints}.
}
    \label{fig:bigtom_desire_prompt}
\end{figure}

\begin{figure}[h]
    \centering
    \scalebox{0.5}{
    \includegraphics{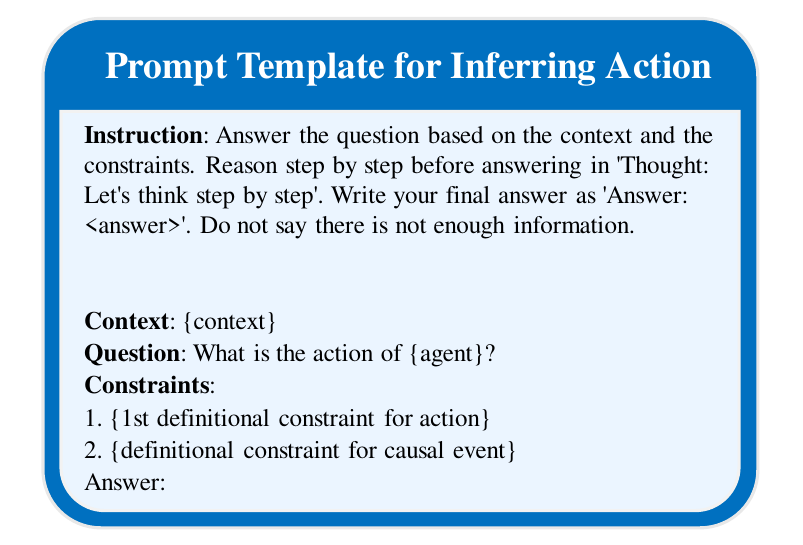}
    }
    \caption{The prompt template for prompting the underlying LLM to infer action for Backward Belief task in BigToM. The constraints shown in the figure are described in Section \ref{sec:complete_constraints}.
}
    \label{fig:bigtom_action_related_prompt}
\end{figure}

The prompt template for inferring percept for Forward Belief and Forward Action tasks is shown in Figure \ref{fig:bigtom_percept_prompt}.
The prompt template for inferring belief for Forward Action tasks is shown in Figure \ref{fig:bigtom_related_belief_prompt}. 
The prompt template for inferring desire for Backward Belief and Forward Action tasks is shown in Figure \ref{fig:bigtom_desire_prompt}. 
The prompt template for inferring action for Backward Belief task is shown in Figure \ref{fig:bigtom_action_related_prompt}. 

\subsubsection{Inferring Queried ToM Dimensions}\label{sec:prompt_templates:bigtom:quried}
\begin{figure}[h]
    \centering
    \scalebox{0.5}{
    \includegraphics{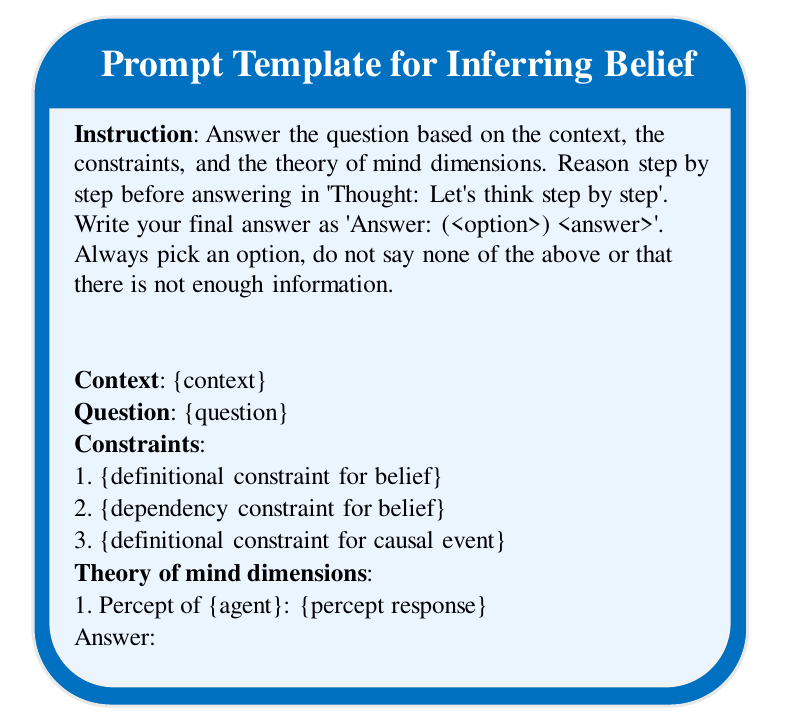}
    }
    \caption{The prompt template for prompting the underlying LLM to infer belief based on the inferred percept for Forward Belief task in BigToM. The constraints shown in the figure are described in Section \ref{sec:complete_constraints}. The ``\{percept response\}’’ refers to the inferred percept.
}
    \label{fig:bigtom_belief_queried_prompt}
\end{figure}

\begin{figure}[h]
    \centering
    \scalebox{0.5}{
    \includegraphics{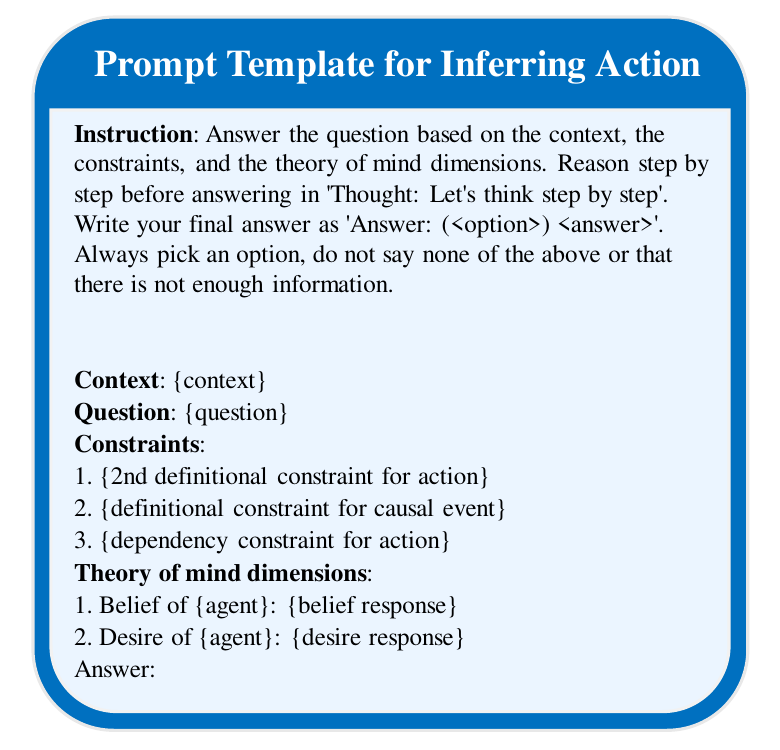}
    }
    \caption{The prompt template for prompting the underlying LLM to infer action based on the inferred belief and desire for Forward Action task in BigToM. The constraints shown in the figure are described in Section \ref{sec:complete_constraints}. The ``\{belief response\}’’ and ``\{desire response\}’’ refer to the inferred belief and desire respectively.
}
    \label{fig:bigtom_action_queried_prompt}
\end{figure}

The prompt template for inferring belief based on the inferred percept for Forward Belief task is shown in Figure \ref{fig:bigtom_belief_queried_prompt}.
The prompt template for inferring action based on the inferred belief and desire for Forward Action task is shown in Figure \ref{fig:bigtom_action_queried_prompt}.

\subsection{FANT{\scriptsize O}M}\label{sec:prompt_templates:fantom}
\begin{figure}[h]
    \centering
    \scalebox{0.5}{
    \includegraphics{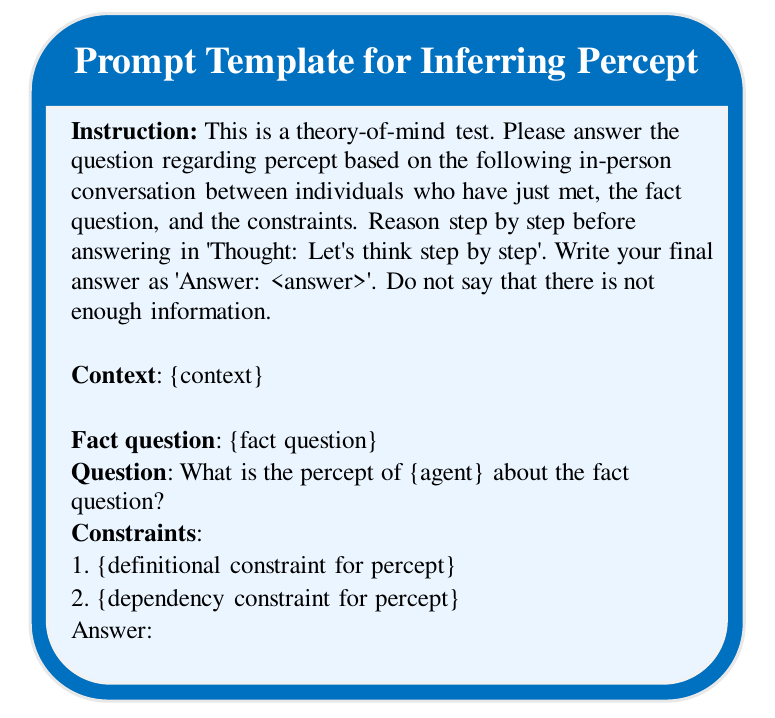}
    }
    \caption{The prompt template for prompting the underlying LLM to infer percept in FANT{\scriptsize O}M. The ``fact question’’ is the question querying the information directly related to the given belief question. We provide an example of fact question and explain how CCoToM reconstructs the fact question from a given question in Section \ref{sec:prompt_templates:fantom}. The constraints shown in the figure are described in Section \ref{sec:complete_constraints}.
}
    \label{fig:fantom_percept_prompt}
\end{figure}

\begin{figure}[h]
    \centering
    \scalebox{0.5}{
    \includegraphics{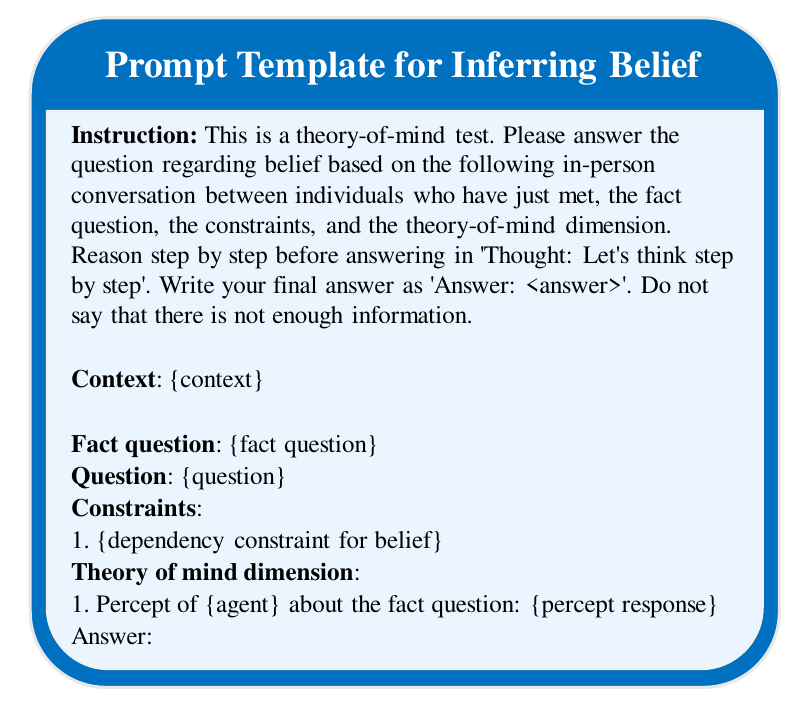}
    }
    \caption{The prompt template for prompting the underlying LLM to infer belief based on the inferred percept for Dist questions in FANT{\scriptsize O}M. The prompt template for  Choice questions is the same except that the last two sentences in the instruction are replaced by ``Write your final answer as `Answer: (<option>) <answer>’. Always pick an option, do not say none of the above or that there is not enough information.’’ The constraint shown in the figure is described in Section \ref{sec:complete_constraints}. The ``\{percept response\}’’ is the inferred percept.
}
    \label{fig:fantom_belief_prompt}
\end{figure}

As mentioned in Section 4.1 of the main content, the questions from FANT{\scriptsize O}M are regarded as questions of Forward Belief task in Section 3.1 of the main content. Hence, CCoToM first prompts the underlying LLM to infer the percept of the target agent. Then, CCoToM prompts the LLM to infer the belief of the target agent based on the inferred percept. The corresponding prompt templates are shown in Figure \ref{fig:fantom_percept_prompt} and Figure \ref{fig:fantom_belief_prompt}.
Given a belief question asking: ``What does Jerry believe is the Sunday tradition of Kali's family?'', the corresponding fact question is ``What is the Sunday tradition of Kali's family?''.
To reconstruct the corresponding fact question from a given belief question, CCoToM first prompts the underlying LLM to answer the following question: “What information about the belief of \{agent\} is queried in the given question?”, then CCoToM prompts the LLM to generate a question asking about the information.

\section{Complete Set of Constraints}\label{sec:complete_constraints}
In this section, we present the complete set of constraints, where the constraints for BigToM and FANT{\scriptsize O}M are shown in Table \ref{tab:bigtom_constraints} and Table \ref{tab:fantom_constraints}.

\begin{table*}[!h]
\caption{The complete set of constraints for BigToM.}
    \label{tab:bigtom_constraints}
    \centering
\scalebox{0.8}{
\begin{tabular}{l|l}
\hline
\multicolumn{1}{c}{\textbf{Constraints}} & \multicolumn{1}{c}{\textbf{Natural Language Instructions}}\\
\hline
Definitional constraint for belief & Belief of \{agent\} is what \{agent\} believes about the state of the environment. \\

Definitional constraint for percept & Percept of \{agent\} is whether or not \{agent\} perceives the causal event. \\

Definitional constraint for causal event & Causal event is the event that changes the state of the environment. \\
1st definitional constraint for action & Action of \{agent\} is what \{agent\} does after the causal event. \\
2nd definitional constraint for action & Action of \{agent\} is what \{agent\} will do after the causal event. \\
Definitional constraint for desire & Desire of \{agent\} is what \{agent\} wants. \\
Dependency constraint for belief & Belief of \{agent\} is determined by the percept of \{agent\}. \\
Dependency constraint for action & Action of \{agent\} is determined by the belief of \{agent\} and the desire of \{agent\}. \\
\hline
\end{tabular}}
\vspace{0.6cm}
\end{table*}

\begin{table*}[!h]
\caption{The complete set of constraints for FANT{\scriptsize O}M.}
\label{tab:fantom_constraints}
\centering
\scalebox{0.8}{
\begin{tabular}{l|l}
\hline
\multicolumn{1}{c}{\textbf{Constraints}} & \multicolumn{1}{c}{\textbf{Natural Language Instructions}}\\
\hline
\multirow{2}{*}{Definitional constraint for percept} & The percept of {agent} about the fact question is whether or not {agent} perceives \\
&  the information about the fact question. \\
\hline
\multirow{4}{*}{Dependency constraint for percept} & If {agent} is absent from the conversation where the information about the fact\\
 & question is shared, {agent} does not perceive the information about the fact question.   \\
 & If {agent} is not absent from the conversation where the information about the fact \\
 & question is shared, {agent} perceives the information about the fact question. \\ 
\hline
\multirow{2}{*}{Dependency constraint for belief} & What {agent} believes about the fact question is determined by the percept of {agent} \\
 & about the fact question. \\
\hline
\end{tabular}
}
\end{table*}

\clearpage
%
%
%
%
%
%
\bibliographystyle{splncs04}
\bibliography{mybib}
\end{document}